\documentclass[10pt,twocolumn,letterpaper]{article}

\usepackage{iccv}
\usepackage{times}
\usepackage{epsfig}
\usepackage{graphicx}
\usepackage{amsmath}
\usepackage{amssymb}
\usepackage{amsmath,subcaption}
\usepackage{epsfig,amssymb,color}

\usepackage{caption}
\usepackage{float}
\usepackage{mathrsfs}
\usepackage{color}
\usepackage{cite}
\usepackage{booktabs}
\usepackage{paralist}
\usepackage{epstopdf}
\usepackage{bbding}
\usepackage[accsupp]{axessibility}
\usepackage{multirow,booktabs,color,soul,threeparttable,rotating}
\usepackage[table]{xcolor}
\def\ourmethod{FaPN}
\usepackage[symbol]{footmisc}
\usepackage[ruled, vlined, linesnumbered]{algorithm2e}

\usepackage[LGR,T1]{fontenc}
\DeclareSymbolFont{upgreek}{LGR}{cmr}{m}{n}
\SetSymbolFont{upgreek}{bold}{LGR}{cmr}{bx}{n}
\DeclareMathSymbol{\uplambda}{\mathord}{upgreek}{`l}

\usepackage{bm}
\usepackage[normalem]{ulem}
\useunder{\uline}{\ul}{}
\usepackage{comment}

\usepackage[pagebackref=true,breaklinks=true,letterpaper=true,colorlinks,linkcolor=red,bookmarks=false]{hyperref}

\iccvfinalcopy % *** Uncomment this line for the final submission

\ificcvfinal\fi

\begin{document}

%%%%%%%%% TITLE
\title{\ourmethod{}: Feature-aligned Pyramid Network for Dense Image Prediction}
 
\author{Shihua Huang \quad Zhichao Lu\quad\quad Ran Cheng\thanks{Corresponding author.}\quad\quad Cheng He\quad\quad\\
{\normalsize Southern University of Science and Technology\thanks{Authors are with Department of Computer Science and Engineering.}}\\
{\tt\small \{shihuahuang95, luzhichaocn, ranchengcn, chenghehust\}@gmail.com}
}

\maketitle
% \thispagestyle{empty}

%%%%%%%%% ABSTRACT
\begin{abstract}
Recent advancements in deep neural networks have made remarkable leap-forwards in dense image prediction.
However, the issue of feature alignment remains as neglected by most existing approaches for simplicity. 
Direct pixel addition between upsampled and local features leads to feature maps with misaligned contexts that, in turn, translate to mis-classifications in prediction, especially on object boundaries. 
In this paper, we propose a feature alignment module that learns transformation offsets of pixels to contextually align upsampled higher-level features; and another feature selection module to emphasize the lower-level features with rich spatial details. 
We then integrate these two modules in a top-down pyramidal architecture and present the Feature-aligned Pyramid Network (\ourmethod{}).
Extensive experimental evaluations on four dense prediction tasks and four datasets have demonstrated the efficacy of \ourmethod{}, yielding an overall improvement of 1.2 - 2.6 points in AP / mIoU over FPN when paired with Faster / Mask R-CNN.
In particular, our \ourmethod{} achieves the state-of-the-art of 56.7\% mIoU on ADE20K when integrated within MaskFormer.
The code is available from \href{https://github.com/EMI-Group/FaPN}{https://github.com/EMI-Group/FaPN}. 
\end{abstract}

\vspace{-0.5cm}
\section{Introduction}

% describe the problem
Dense prediction is a collection of computer vision tasks that aim at labeling every pixel in an image with a pre-defined class. 
It plays a fundamental role in scene understanding and is of great importance to real-world applications, such as autonomous driving~\cite{Chen_2016_CVPR}, medical imaging~\cite{wang2017chestx}, augmented reality~\cite{alhaija2018augmented}, \etc 
The modern solutions for these tasks are built upon Convolutional Neural Networks (CNNs). 
With the recent advancements in CNN architectures, a steady stream of promising empirical leap-forwards was reported across a wide range of dense prediction tasks, including object detection~\cite{ren2015faster, liu2016ssd, redmon2016you}, semantic segmentation~\cite{long2015fully, chen2017deeplab}, instance segmentation~\cite{he2017mask,Liu_2018_CVPR}, and panoptic segmentation~\cite{kirillov2019panoptic1, kirillov2019panoptic}, to name a few.

\begin{figure}
    \centering
    \begin{subfigure}{0.4\textwidth}
    % \begin{center}
    \hspace{.7em}\includegraphics[width=\textwidth]{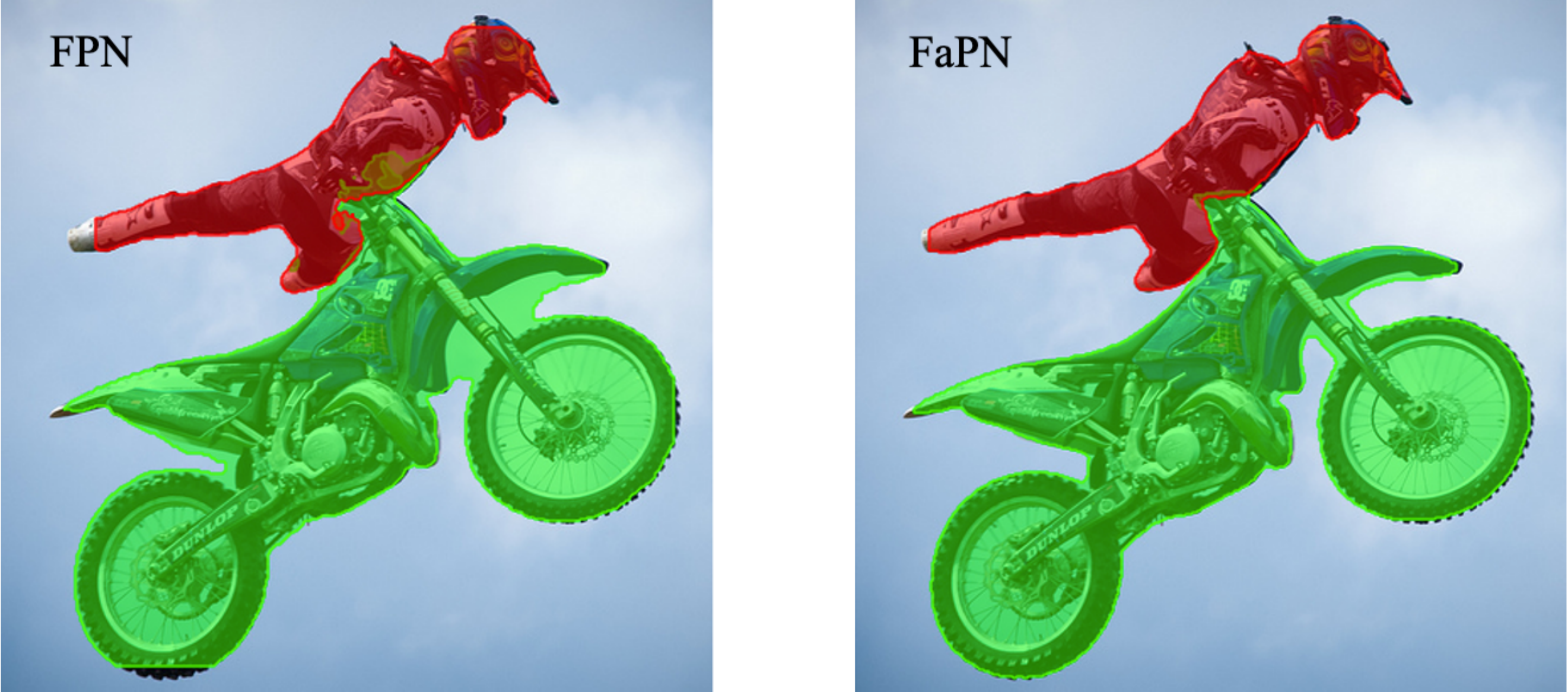}
    % \end{center}
    \end{subfigure}\\\vspace{-1em}
    \centering
    \begin{subfigure}{0.46\textwidth}
    \centering
    \vspace{0.5cm}
    \includegraphics[width=\textwidth{}]{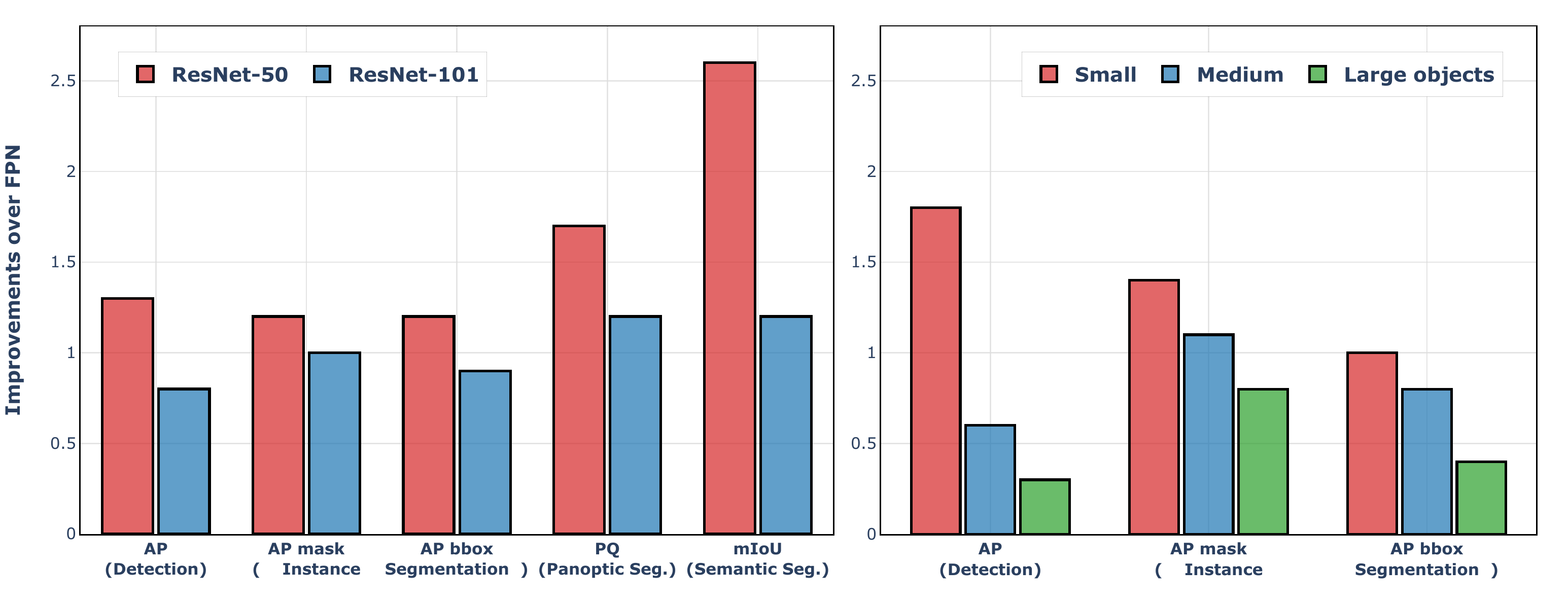}
    \end{subfigure} 
    \caption{\textbf{Comparisons between FPN and \ourmethod{}}:
    (\emph{Top row}) Qualitatively, \ourmethod{} significantly improves the performance on object boundaries as opposed to its counterpart, \ie FPN~\cite{lin2017feature}.
    (\emph{Bottom row}) Quantitatively, \ourmethod{}'s improvements over FPN are consistent across different tasks, backbones, and object scales. Best view in color.
    }
    \label{fig:font_com}
     \vspace{-0.5cm}
\end{figure}

\begin{comment}
\begin{figure}
    \begin{subfigure}{0.46\textwidth}
    \begin{center}
    \includegraphics[width=0.98\linewidth]{figures/PointRend_w_vs_wo_FaPN.pdf}
    \end{center}
    \end{subfigure} \\
    \centering
    \begin{subfigure}{0.46\textwidth}
    \centering
    \vspace{0.5cm}
    \includegraphics[width=\textwidth{}]{figures/preview_improvement.pdf}
    \end{subfigure} 
    \caption{\textbf{Comparisons between FPN and \ourmethod{}}. \ourmethod{} (\emph{Feature-aligned Pyramid Network}) learns a set of transformation offsets to align the bottom-up details and the upsampled contexts. Qualitatively, \ourmethod{} (\emph{Upper-Right}) significantly improves the performance on object boundaries as opposed to its counterpart, i.e. FPN~\cite{lin2017feature} (\emph{Upper-Left}). Quantitatively, \ourmethod{}'s improvements over FPN are consistent across different tasks, backbones, and object scales (\emph{Bottom}). Best view in color.}
     \label{fig:font_com}
     \vspace{-0.3cm}
\end{figure}
\end{comment}

\begin{figure*}
\begin{center}
\includegraphics[width=.95\linewidth]{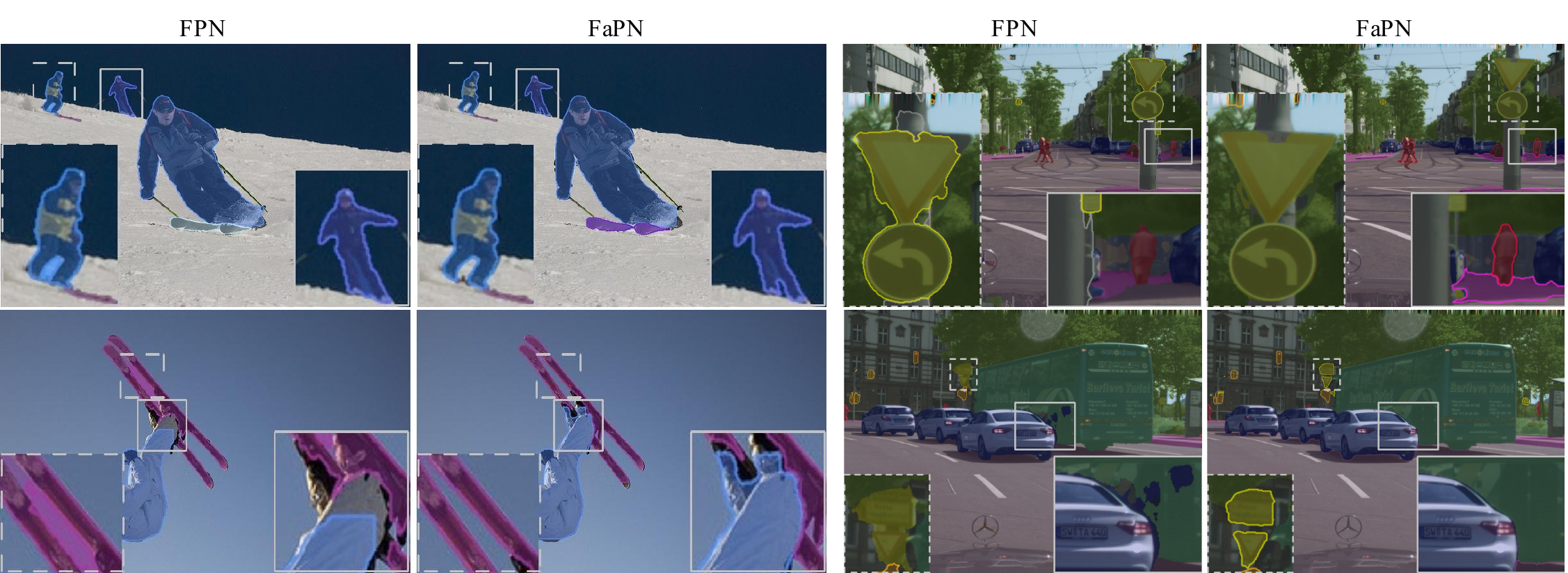}
\end{center}
\vspace{-0.2cm}
   \caption{\textbf{Example pairs of results from FPN~\cite{lin2017feature} and our \ourmethod{}}. Both methods are implemented in Mask R-CNN~\cite{he2017mask} with ResNet50~\cite{resnet} being the backbone and PointRend~\cite{kirillov2020pointrend} as the mask head. Qualitatively, \ourmethod{} significantly improves the performance on object boundaries. Images are randomly chosen from \cite{lin2014microsoft} and \cite{cordts2016cityscapes} for instance (\emph{left}) and semantic (\emph{right}) segmentation, respectively. More visualization examples are available in the supplementary materials.}
\label{fig:preview_results}
\vspace{-0.1cm}
\end{figure*}

Dense prediction requires both rich spatial details for object location and strong semantics for object classification, which most likely reside at different resolution / scale levels~\cite{long2015fully}. 
How to effectively generate a hierarchy of features at different scales becomes one of the key barriers to overcome in handling dense prediction tasks~\cite{lin2017feature}.
{Broadly speaking}, there are two common practices to address this issue. 
The first kind uses atrous convolutions with different atrous rates to effectively capture long-range information (\ie semantic context) without reducing spatial resolution~\cite{chen2017deeplab}. % 
The other kind builds a top-down feature pyramid based on the default bottom-top pathway of a ConvNet \cite{badrinarayanan2017segnet}.
More specifically, the (higher-level) spatially coarser feature maps are upsampled before merging with the corresponding feature maps from the bottom-up path-way. 
However, there are inaccurate correspondences (\ie \emph{feature misalignment}) between the bottom-up and upsampled features owing to the non-learnable nature of the commonly-used upsampling operations (\eg nearest neighbor) and the repeated applications of downsampling and upsampling.
The misaligned features, in turn, adversely affects the learning in the subsequent layers, resulting in mis-classifications in the final predictions, especially around the object boundaries.
To address the aforementioned issue, we propose a feature alignment module that learns to align the upsampled feature maps to a set of reference feature maps by adjusting each sampling location in a convolutional kernel with a learned offset.
We further propose a feature selection module to adaptively emphasize the bottom-up feature maps containing excessive spatial details for accurate locating.  
We then integrate these two modules in a top-down pyramidal architecture and propose the \emph{Feature-aligned Pyramid Network (\ourmethod{})}. 

Conceptually, \ourmethod{} can be easily incorporated to existing bottom-up ConvNet backbones \cite{resnet,muxconv,nsganetv1,nat} to generate a pyramid of features at multiple scales~\cite{lin2017feature}. We implement \ourmethod{} in modern dense prediction frameworks (Faster R-CNN~\cite{ren2015faster}, Mask R-CNN~\cite{he2017mask}, PointRend~\cite{kirillov2020pointrend}, MaskFormer~\cite{cheng2021per},  PanopticFPN~\cite{kirillov2019panoptic}, and PanopticFCN~\cite{li2021fully}), and demonstrate its efficacy on object detection, semantic, instance and panoptic segmentation. Extensive evaluations on multiple challenging datasets suggest that \ourmethod{} leads to a significant improvement in dense prediction performance, especially for small objects and on object boundaries. Moreover, \ourmethod{} can also be easily extended to real-time semantic segmentation by pairing it with a lightweight bottom-up backbone \cite{resnet,nsganet,nsganetv2}. Without bells and whistles, \ourmethod{} achieves favorable performance against existing dedicated real-time methods. Our key contributions are:

\vspace{2pt}
\noindent\textbf{--} We first develop (i) a feature alignment module that learns transformation offsets of pixels to contextually align upsampled (higher-level) features; and (ii) another feature selection module to emphasize (lower-level) features with rich spatial details. 

\vspace{2pt}
\noindent\textbf{--} With the integration of these two contributions, we present, \emph{Feature-aligned Pyramid Network (\ourmethod{})}, an enhanced drop-in replacement of FPN~\cite{lin2017feature}, for generating multi-scale features. 

\vspace{2pt}
\noindent\textbf{--} We present a thorough experimental evaluation demonstrating the efficacy and value of each component of \ourmethod{} across \emph{four} dense prediction tasks, including object detection, semantic, instance, and panoptic segmentation on three benchmark datasets, including MS COCO \cite{lin2014microsoft}, Cityscapes \cite{cordts2016cityscapes}, COCO-Stuff-10K \cite{caesar2018coco}. 

\vspace{2pt}
\noindent\textbf{--} Empirically, we demonstrate that our \ourmethod{} leads to a significant improvement of \textbf{1.2\% - 2.6\%} in performance (AP / mIoU) over the original FPN~\cite{lin2017feature}. Furthermore, our \ourmethod{} achieves the state-of-the-art of 56.7\% mIoU on ADE20K when integrated within MaskFormer \cite{cheng2021per}.

\section{Related Work}  \label{related_work}

\noindent\textbf{Feature Pyramid Network Backbone:}
The existing dense image prediction methods can be broadly divided into two groups. 
The first group utilizes atrous convolutions to enlarge the receptive field of convolutional filters for capturing long-range information without reducing resolutions spatially. 
DeepLab~\cite{chen2017deeplab} is one of the earliest method that adopt atrous convolution for semantic segmentation. 
It introduced an Atrous Spatial Pyramid Pooling module (ASPP) comprised of atrous convolutions with different atrous rates to aggregate multi-scale context from high-resolution feature maps.
Building upon ASPP, a family of methods~\cite{chen2017deeplab, chen2017rethinking, chen2018encoder} were developed. 
However, the lack of the ability to generate feature maps at multiple scales restricts the application of this type of methods to other dense prediction tasks beyond semantic segmentation.
The second group of methods focuses on building an encoder-decoder network, \ie bottom-up and top-down pathways.
The top-down pathway is used to back-propagate the high-level semantic context into the low-level features via a step-by-step upsampling.
There is a plethora of encoder-decoder methods~\cite{noh2015learning, fu2017dssd, lin2017feature, he2017mask, kirillov2019panoptic, wang2020solo, wang2020solov2} proposed for different dense image prediction tasks. 
DeconvNet~\cite{noh2015learning} is one of the earliest works that proposed to use upsample operations with learnable parameters, \ie deconvolution. 
DSSD~\cite{fu2017dssd} and FPN~\cite{lin2017feature} are the extensions of SSD~\cite{liu2016ssd} and Faster R-CNN~\cite{ren2015faster} respectively for object detection.
Mask R-CNN~\cite{he2017mask} and SOLOs~\cite{wang2020solo, wang2020solov2} are used for real-time instance segmentation. 
Moreover, Kirillov \etal propose the Panoptic FPN~\cite{kirillov2019panoptic} for panoptic segmentation. 

\vspace{1em}
\noindent\textbf{Feature Alignment:}
In case of the increasing loss of boundary detail with the step-by-step downsampling, SegNet\cite{badrinarayanan2017segnet} stores the max-pooling indices in its encoder and upsamples feature maps in the decoder with the corresponding stored max-pooling indices. 
Instead of memorizing the spatial information in the encoder previously as SegNet, GUN~\cite{mazzini2018guided} tries to learn the guidance offsets before upsampling in the decoder and then upsamples feature maps following those offsets.
To solve the misalignment between extracted features and the RoI caused by the quantizations in RoIPool, RoIAlign~\cite{he2017mask} avoids any quantizations and computes the values for each RoI with linear interpolation.
To establish accurate correspondences among multiple frames given a large motion for video restoration, TDAN~\cite{tian2020tdan} and EDVR~\cite{wang2019edvr} achieve implicit motion compensation with by deformable convolution~\cite{dai2017deformable} at the feature level. 
AlignSeg~\cite{huang2020alignseg} and SFNet~\cite{li2020semantic} are two concurrent works that share a similar motivation as ours and both are flow-based alignment methods. 
In particular, AlignSeg proposes a two-branched bottom-up network and uses two types of alignment modules to alleviate the feature misalignment before feature aggregation. 
In contrast, we propose to construct a top-down pathway based on the bottom-up network and align features from the coarsest resolution (top) to the finest resolution (bottom) in a progressive way.
Specifically, we only align $2\times$ upsampled features to their corresponding bottom-up features, while AlignSeg tries to align diversely scaled features (\ie upsampled from $1/4$, $1/8$, and even $1/16$) directly which are difficult and may not always be feasible.

\begin{figure}[t]
\begin{center}
\includegraphics[width=0.99\linewidth]{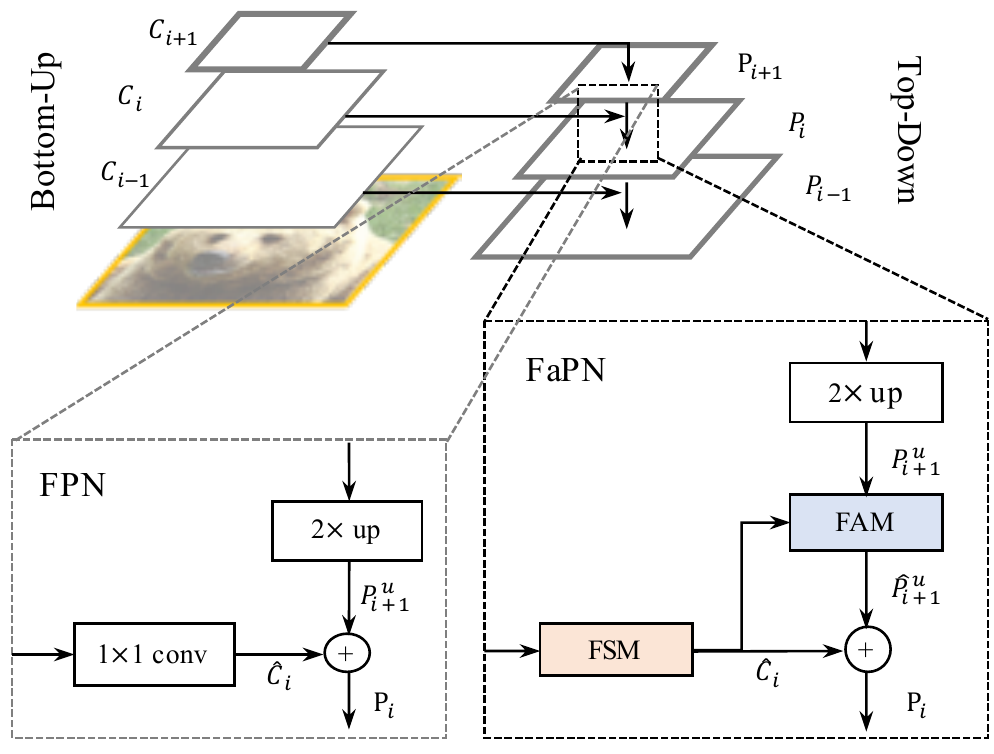}
\end{center}
\vspace{-0.2cm}
   \caption{\textbf{Overview comparison between FPN and \ourmethod{}}. Details of the {FAM} and {FSM} components are provided in Figure~\ref{fig:fam} and Figure~\ref{fig:fsm}, respectively. 
   }
\label{fig:fpn_vs_fan}
\vspace{-0.2cm}
\end{figure}

\section{Feature-aligned Pyramid Network}\label{method}

% General framework 
In this section, we present the general framework of our method, comprised of a {Feature Selection Module} (FSM) and a {Feature Alignment Module} (FAM), as shown in Figure~\ref{fig:fpn_vs_fan} (\emph{right}).
Specifically, we define the output of the $i$-th stage of the bottom-up network as $\mathbf{C}_i$, which has stride of $2^i$ pixels with respect to the input image, \ie $\mathbf{C}_{i} \in \mathbb{R}^{\frac{H}{2^i} \times \frac{W}{2^i}}$, where $H\times W$ is the size of the input image. \textcolor{black}{And we denote ($\frac{H}{2^i}, \frac{W}{2^i}$) by ($H_{i}, W_{i})$} for brevity. We use $\mathbf{\hat{C}}_i$ to denote the output of a FSM layer given the input of $\mathbf{C}_i$.
Also, the output after the $i$-th feature fusion in the top-down pathway is defined as $\mathbf{P}_{i}$, and its upsampled and aligned features to $\mathbf{C}_{i-1}$ as $\mathbf{P}_{i}^{u}$ and $\mathbf{\hat{P}}_{i}^{u}$, respectively. 

\subsection{Feature Alignment Module}

\begin{figure}[!ht]
\begin{center}
\includegraphics[width=1\linewidth]{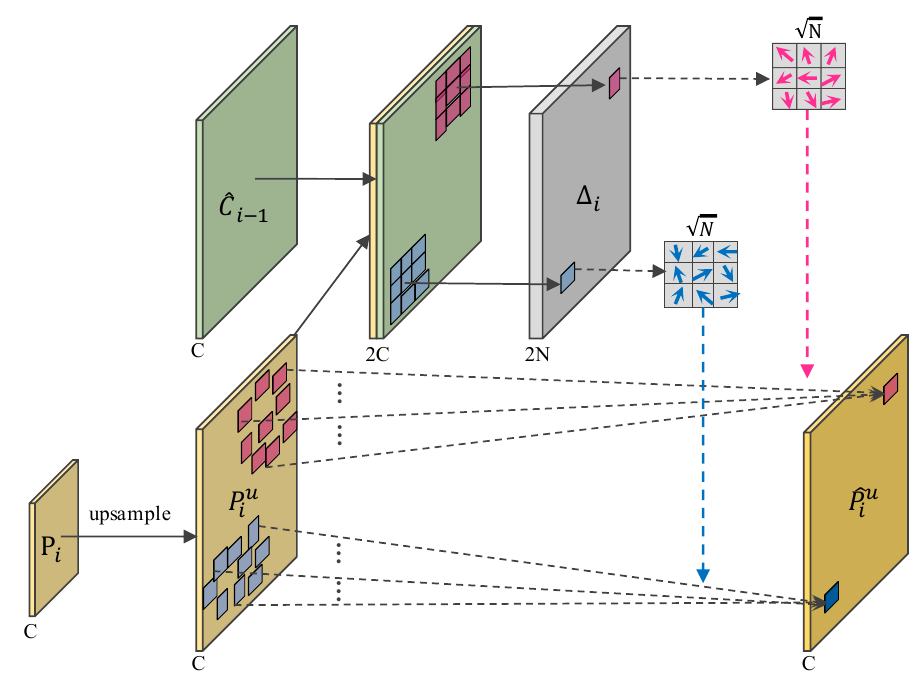}
\end{center}
\vspace{-2em}
\caption{\textbf{Feature Alignment Module}. The offset fields have the same spatial resolution with the input and 2$N$ channels corresponding to $N$ 2D offsets. Specifically, $N$ denotes a convolutional kernel of $N$ sample locations, \eg $N$ is equal to 9 for a $3\times 3$ conv, and each value in the $n$-th offset filed is the horizontal or vertical offset for the $n$-th sample point.}
\label{fig:fam}
\vspace{-1em}
\end{figure}

Due to the recursive use of downsampling operations, there are foreseeable spatial misalignment between the upsampled feature maps $\mathbf{P}^{u}_{i}$ and the corresponding bottom-up feature maps $\mathbf{C}_{i-1}$.
Thus, the feature fusion by either element-wise addition or channel-wise concatenation would harm the prediction around object boundaries.
Prior to feature aggregation, aligning $\mathbf{P}^{u}_{i}$ to its reference $\mathbf{\hat{C}}_{i-1}$ is essential, \ie adjusting $\mathbf{P}^{u}_{i}$ accordingly to the spatial location information provided by the $\mathbf{\hat{C}}_{i-1}$.
In this work, the spatial location information is presented by 2D feature maps, where each offset value can be viewed as the shifted distances in 2D space between each point in $\mathbf{P}^{u}_{i}$ and its \textcolor{black}{corresponding point} in $\mathbf{\hat{C}}_{i-1}$. 
As illustrated by Figure~\ref{fig:fam}, the feature alignment can be mathematically formulated as: 
\begin{equation}\label{eq:general_align}
\begin{split}
    \mathbf{\hat{P}}^{u}_{i} & = f_{a}\big(\mathbf{P}^{u}_{i}, \mathbf{\Delta}_{i}\big), \\
    \mathbf{\Delta}_{i} & = f_{o}\big([\mathbf{\hat{C}}_{i-1}, \mathbf{P}_{i}^{u}]\big),
\end{split}
\end{equation} 
where $[\mathbf{\hat{C}}_{i-1}, \mathbf{P}_{i}^{u}]$ is the concatenation of $\mathbf{\hat{C}}_{i-1}$ and $\mathbf{P}_{i}^{u}$ which provides  spatial difference between the upsampled and corresponding bottom-up features. $f_{o}(\cdot)$ and $f_{a}(\cdot)$ denote the functions for learning offsets ($\mathbf{\Delta_{i}}$) from the spatial differences and aligning feature with the learned offsets, respectively. 
In this work, $f_{a}(\cdot)$ and $f_{o}(\cdot)$ are implemented using deformable convolutions~\cite{dai2017deformable, zhu2019deformable}, followed by activation and standard convolutions of the same kernel size. 

Here, we briefly review the deformable convolution \cite{dai2017deformable}, and then explain why it can be used as our feature alignment function and provide some important implementation details.  
We first define an input feature map $\mathbf{c}_{i} \in \mathbb{R}^{H_{i} \times W_{i}}$ and a $k\times k$ conv layer.
Then, the output feature at any position $\hat{x}_{\mathbf{p}}$\footnote{where $\mathbf{p} \in \{(0, 0), (1, 0), (0, 1), \ldots, (H_i-1, W_i-1)\}$} after the convolutional kernel can be obtained by
\begin{eqnarray}\label{eq:conv}
    \hat{x}_{\mathbf{p}} = \sum_{n=1}^{N}w_n\cdot x_{\mathbf{p}+\mathbf{p}_{n}}, 
\end{eqnarray}
where $N$ is the size of the $k\times k$ convolutional layer (\ie $N = k\times k$),  $w_n$ and $\mathbf{p}_n$ $\in \{(-\lfloor \frac{k}{2} \rfloor, -\lfloor \frac{k}{2} \rfloor), (-\lfloor \frac{k}{2} \rfloor, 0), \ldots, (\lfloor \frac{k}{2} \rfloor, \lfloor \frac{k}{2} \rfloor)\}$ refer to the weight and the pre-specified offset for the $n$-th convolutional sample location, respectively. 
In addition to the pre-specified offsets, the deformable convolution tries to learn additional offsets $\{\Delta\mathbf{p}_1, \Delta\mathbf{p}_2, ..., \Delta\mathbf{p}_N\}$ adaptively for different sample locations, and Equation~(\ref{eq:conv}) can be reformulated as
\begin{eqnarray}\label{eq:dcn}
    \hat{x}_{\mathbf{p}} = \sum_{n=1}^{N}w_n\cdot x_{\mathbf{p}+\mathbf{p}_{n} + \Delta\mathbf{p}_{n}}, 
\end{eqnarray}
where each $\Delta\mathbf{p}_{n}$ is a tuple $(h, w)$, with $h \in (-H_{i}, H_{i})$ and $w \in (-W_{i}, W_{i})$. 

When we apply the deformable convolution over the $\mathbf{P}^{u}_{i}$ and take the concatenation of $\mathbf{\hat{C}}_{i-1}$ and $\mathbf{P}_{i}^{u}$ as the reference (\ie offset fields $\mathbf{\Delta}_{i} = f_{o}\big([\mathbf{\hat{C}}_{i-1}, \mathbf{P}_{i}^{u}]\big)$), the deformable convolution can adjust its convolutional sample locations following the offsets following Equation (\ref{eq:general_align})\footnote{Following the convention of the deformable convolution, this study adopts $3\times3$ as the kernal size for $f_{a}(\cdot)$ and $f_{o}(\cdot)$.}, \ie aligning $\mathbf{P}^{u}_{i}$ according to the spatial distance between $\mathbf{\hat{C}}_{i-1}$ and $\mathbf{P}^{u}_{i}$. 

\subsection{Feature Selection Module}
Prior to channel reduction for detailed features, it is vital to emphasize the important feature maps that contain excessive spatial details for accurate allocations while suppressing redundant feature maps. 
Instead of simply using a $1\times1$ convolution~\cite{lin2017feature}, we propose a feature selection module (FSM) to explicitly model the importance of feature maps and re-calibrate them accordingly.

\begin{figure}[!ht]
\begin{center}
\includegraphics[width=1\linewidth]{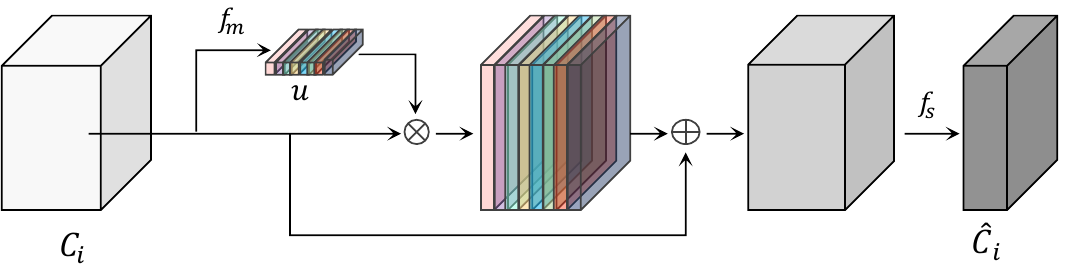}
\end{center}
\vspace{-1em}
   \caption{\textbf{Feature Selection Module}. $\mathbf{C}_{i}$ = [$\mathbf{c}_1$, $\mathbf{c}_2$, \ldots, $\mathbf{c}_D$] and $\mathbf{\hat{C}}_{i}$ = [$\mathbf{\hat{c}}_1$, $\mathbf{\hat{c}}_2$, \ldots, $\mathbf{\hat{c}}_{D'}$] refer to the input and output feature maps respectively, where $\mathbf{c_d}$ and $\mathbf{\hat{c}_{d'}}$ $\in \mathbb{R}^{H_{i}\times W_{i}}$, $D$ and $D'$ denote the input and output channels, respectively. $\mathbf{u}$ = [$u_{1}$, $u_{2}$, \ldots, $u_{D}$] is the feature importance vector, where $u_{d}$ represents the importance of the $d$-th input feature map. \textcolor{black}{$f_m$ and $f_s$ represent the feature importance modeling and feature selection layer, respectively. See text for details.}}
\label{fig:fsm}
\end{figure}

The general dataflow of the proposed FSM is presented in Figure~\ref{fig:fsm}.
To begin with, the global information $\mathbf{z}_{i}$ of each input feature map $\mathbf{c}_{i}$ is extracted by a global average pooling operation, while a feature importance modeling layer $f_m(\cdot)$ (\ie a $1\times1$ conv layer followed by a sigmoid activation function) learns to use such information for modeling the importance of each feature map and outputs an importance vector $\mathbf{u}$.  
Next, the original input feature maps are scaled with the importance vector, and then the scaled feature maps are added to the original feature maps, referred as rescaled feature maps.
Finally, a feature selection layer ${f}_{s}(\cdot)$ (\ie a $1\times1$ conv layer for efficiency) is introduced over the rescaled feature maps, which is used to selectively maintain important feature maps and drop useless feature maps for channel reduction.
Overall, the process of FSM can be formulated as
\begin{equation}\label{eq:fsm}
    \begin{split}
    \mathbf{\hat{C}}_i &= f_{s}(\mathbf{C}_{i} + \mathbf{u} \ast \mathbf{C}_{i}), \\ 
    \mathbf{u} &= f_{m}(\mathbf{z}),
    \end{split}
\end{equation}
where $\mathbf{z}$ = [$z_{1}$, $z_{2}$, \ldots, $z_{D}$] and is calculated by
\begin{eqnarray}\label{eq:global_info}
    z_{d} = \frac{1}{H_{i}\times W_{i}}\sum_{h=1}^{H_{i}}\sum_{w=1}^{W_{i}}c_{d}(h, w).
\end{eqnarray}

It is worth mentioning that the design of our FSM is motivated by the squeeze-and-excitation (SE)~\cite{hu2018squeeze}. The main difference lies in the additional skip connection introduced between the input and scaled feature maps (Figure~\ref{fig:fsm}). Empirically, we find that lower bounding the scaled feature (through the skip connection) is essential, which avoids any particular channel responses to be over-amplified or -suppressed. Conceptually, both of these two modules learn to adaptively re-calibrate channel-wise responses by channel attention. However, SE is conventionally used in the backbone for enhancing feature extraction, while FSM is used in the neck (\ie top-down pathway) for enhancing multi-scale feature aggregation. Additionally, the selected/scaled features from FSM are also supplied as references to FAM for learning alignment offsets.

\section{Experiments}   \label{experiments}

In this section, we first briefly introduce the benchmark datasets studied in this work, followed by the implementation and training details. We then evaluate the performance of the proposed \ourmethod{} on four dense image prediction tasks, including object detection, semantic, instance and panoptic segmentation. Ablation studies demonstrating the effectiveness of each component in \ourmethod{} are also provided. Moreover, we incorporate our proposed \ourmethod{} with lightweight backbones and evaluate its efficacy under real-time settings. 

\vspace{3pt}
\noindent\textbf{Datasets:} 
We consider four widely-used benchmark datasets to evaluate our method, including MS COCO~\cite{lin2014microsoft} for object detection, instance and panoptic segmentation;  Cityscapes~\cite{cordts2016cityscapes}, COCO-Stuff-10K~\cite{caesar2018coco} and ADE20K \cite{zhou2017scene} for semantic segmentation.

MS COCO consists of more than 100K images containing diverse objects and annotations, including both bounding boxes and segmentation masks. 
We use the \emph{train2017} set (around 118K images) for training and report results on the \emph{val2017} set (5K images) for comparison. 
For both object detection and instance segmentation tasks, there are 80 categories; and for panoptic segmentation task, there are 80 things and 53 stuff classes annotated. 

Cityscapes is a large-scale dataset for semantic understanding of urban street scenes. 
It is split into training, validation and test sets, with 2975, 500 and 1525 images, respectively. 
The annotation includes 30 classes, 19 of which are used for semantic segmentation task. 
The images in this dataset have a higher and unified resolution of $1024\times2048$, which poses stiff challenges to the task of real-time semantic segmentation. 
For the experiments shown in this part, we only use images with fine annotations to train and validate our proposed method. 

COCO-Stuff-10K contains a subset of 10K images from the COCO dataset~\cite{lin2014microsoft} with dense stuff annotations. 
It is a challenging dataset for semantic segmentation as it has 182 categories (91 thing classes plus 91 stuff classes). 
In this work, we follow the official split -- 9K images for training and 1K images for test.

ADE20K is a challenging scene parsing dataset that contains 20k images for training and 2k images for validation. Images in the dataset are densely labeled as hundreds of classes. In this work, only 150 semantic categories are selected to be included in the evaluation.

\vspace{3pt}
\noindent\textbf{Implementation details:} 
Following the original work of FPN~\cite{lin2017feature}, we use ResNets~\cite{he2016deep} pre-trained on ImageNet~\cite{deng2009imagenet} as the backbone ConvNets for the bottom-up pathway. 
We then replace the FPN with our proposed \ourmethod{} as the top-down pathway network. 
Next, we connect the feature pyramid with the Faster R-CNN detector~\cite{ren2015faster} for object detection, and Mask R-CNN (with PointRend masking head ~\cite{kirillov2020pointrend}) for segmentation tasks.

For performance evaluation, the Average Precision (AP) is used as the primary metric for both object detection and instance segmentation. 
We evaluate AP on small, medium and large objects, \ie AP$_s$, AP$_m$, and AP$_l$.
Note that AP$^{bb}$ and AP$^{mask}$ denote AP for bounding box and segmentation mask, respectively. 
The mean Intersection-over-Union (mIoU) and the Panoptic Quality (PQ) are two primary metrics used for semantic and panoptic segmentation, respectively. 
Additionally, we also use PQ$^{St}$ and PQ$^{Th}$ metrics to evaluate stuff and thing performances separately for panoptic segmentation. 

\subsection{Ablation Study}
We first breakdown the individual impacts of the two components introduced in \ourmethod{}, \ie the feature alignment and selection modules. Using ResNet50 as the bottom-up backbone, we evaluate on Cityscapes for semantic segmentation. 
Table~\ref{tab:ablations} shows the improvement in accuracy along with the complexity overheads measured in \#Params. 

\begin{table}[!ht]
\centering
\caption{\textbf{Ablative Analysis:} \textcolor{black}{Comparing the performance of our \ourmethod{} with other variants on Cityscapes for semantic segmentation. $\dagger$ denotes placing FAM after feature fusion. ``deconv'' refers to the deconvolution which is a learnable upsample operation. The relative \textcolor{red}{improvements}/\textcolor{blue}{overheads} are shown in parenthesis.}\label{tab:ablations}}
% \vspace{-3.0mm}
\resizebox{.42\textwidth}{!}{%
\begin{tabular}{@{}|l|c|rr|@{}}
\hline
method & backbone & \#Params (M) & mIoU (\%) \\ \hline\hline
FPN & R50 & 28.6 (\textcolor{blue}{+4.5}) & 77.4 (\textcolor{red}{+2.6})\\
FPN + extra 3$\times$3 conv. & R50 & 33.4 (\textcolor{red}{-0.3}) & 77.5 (\textcolor{red}{+2.5})\\
FPN & R101 & 47.6 (\textcolor{red}{-14.5}) & 78.9 (\textcolor{red}{+1.1})\\
\hline\hline
FPN + FAM & R50 & 31.7 (\textcolor{blue}{+1.4}) & 79.7 (\textcolor{red}{+0.3})\\ 
FPN + FAM + SE & R50 & 33.1 (+0.0) & 78.8 (\textcolor{red}{+1.2})\\     % SE module 
\rowcolor{lightgray!60}
FPN + FAM + FSM (\ourmethod{}) & R50 & 33.1 (+0.0) & \textbf{80.0} (+0.0)\\ 
\hline\hline
FPN + deconv + FSM & R50 & 32.7 (\textcolor{blue}{+0.4}) & 76.7 (\textcolor{red}{+3.3})\\     % FSM + Deconv
FPN + FAM$^{\dagger}$ + FSM & R50 & 32.7 (\textcolor{blue}{+0.4}) & 79.3 (\textcolor{red}{+0.7})\\\hline    % Variation of FaPN
\end{tabular}%
}
\end{table}

Evidently, with marginal increments in model size, our proposed feature alignment module alone significantly boosts the performance of the original FPN \cite{lin2017feature}, yielding an improvement of \textbf{2.3 points} in mIoU. In particular, our method (\emph{80.0@33.1M}) is significantly more effective than naively expanding either i) the \#Params of FPN by extra 3$\times$3 conv. (\emph{77.5@33.4M}) or ii) the capacity of the backbone from R50 to R101 (\emph{78.9@47.6M}).
Empirically, we observe that a naive application of SE~\cite{hu2018squeeze} (for feature selection) adversely affects the performance, while our proposed FSM provides a further boost in mIoU. 

Recall that the \emph{misalignment} in this work refers to the spatial misalignment of features induced during the aggregation of multi-resolution feature maps (i.e., top-down pathway in FPN), particularly around object boundaries. One plausible cause relates to the non-learnable nature of commonly-used upsampling operations (e.g., bilinear). However, simply swapping it to a learnable operation (e.g., deconvolution) is insufficient, suggesting the need of better engineered methods. This reinforces the motivation of this work.
Instead of performing the feature alignment before feature fusion, we place our FAM after feature fusion, in which our FAM learns the offsets from the fused features instead. Although this variation performs better than all other variants, it is still substantially worse than the proposed \ourmethod{}, which reiterate the necessity of feature alignment before fusion. 

\begin{figure}[!ht]  
	\centering
	\includegraphics[width=0.49\textwidth]{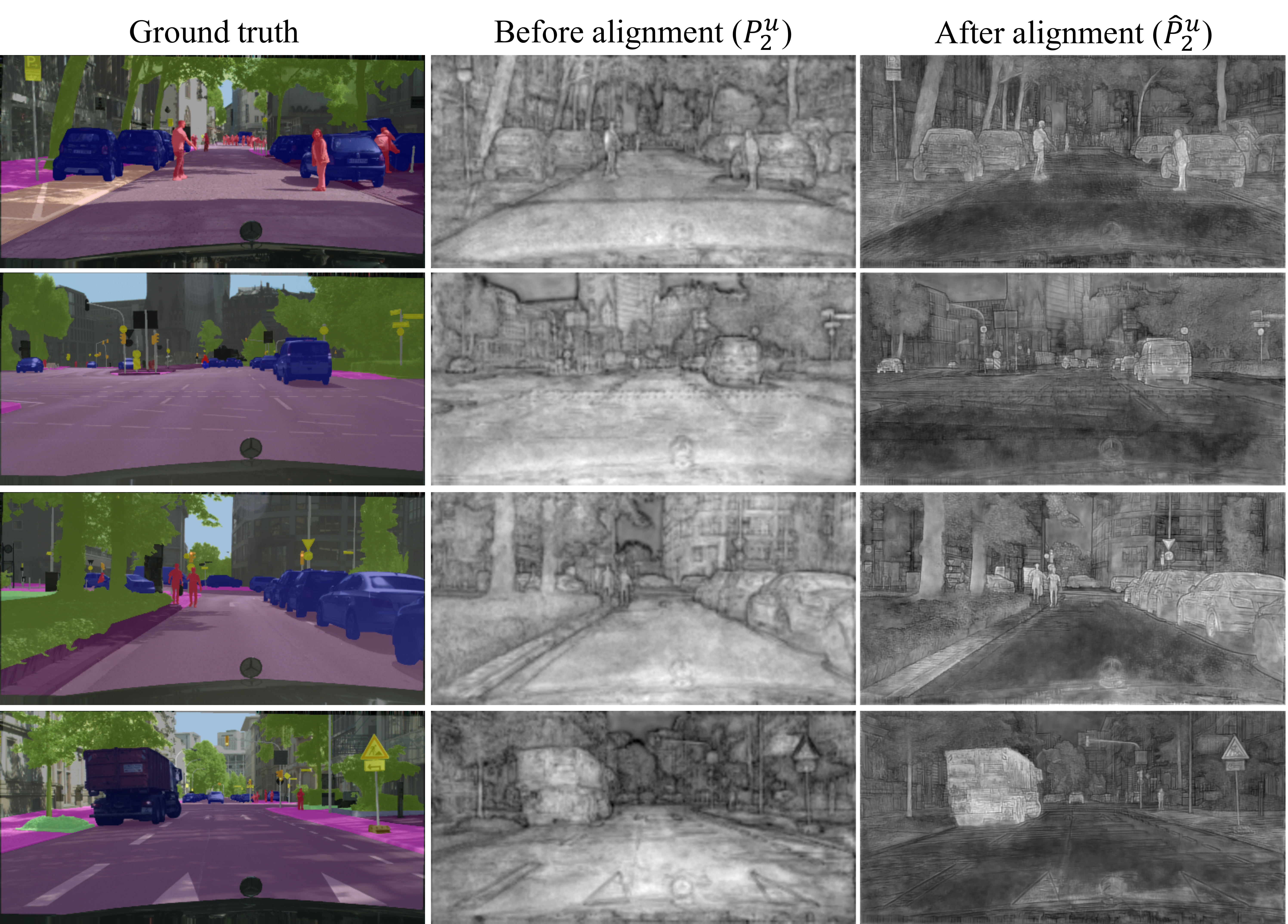}
	\vspace{-0.3cm}
	\caption{Visualization of the input (upsampled features) to and the output (aligned features) from our FAM. Zoom in for better details. \label{fig:visual_feat}}
\end{figure}

\textcolor{black}{
\subsection{Boundary Prediction Analysis}
We provide the mIoU over the boundary pixels\footnote{we consider $n$ pixels around the outline of each object to be boundary pixels, where $n$ can be one of [3, 5, 8, 12].} in Table~\ref{tab:boundary}. Evidently, our method achieves a substantially better segmentation performance than FPN on boundaries. Moreover, we visualize the input (upsampled features $P^{u}_{2}$) to and the output (aligned features $\hat{P}^{u}_{2}$) from the last feature alignment module in FaPN-R50 (Figure~\ref{fig:visual_feat}) to perceive the alignment corrections made by our FAM. In contrast to the raw upsampled features (before FAM) which are noisy and fluctuating, the aligned features are smooth and containing more precise object boundaries. Both the quantitative evaluation and qualitative observation are consistent and suggest that \ourmethod{} leads to better predictions on the boundaries.} More visualizations are provided in Figure~\ref{fig:preview_results}.

\begin{table}[!ht]
\centering
\caption{\textcolor{black}{\textbf{Segmentation Performance around Boundaries:} Comparing the performance of our \ourmethod{} with the original FPN \cite{lin2017feature} in terms of mIoU over boundary pixels on Cityscape \emph{val} with different thresholds on boundary pixels.} \label{tab:boundary}} 
\vspace{-1em}
\resizebox{.48\textwidth}{!}{%
\begin{tabular}{@{}|l|c|ccccc|@{}}
\hline
method & backbone  & 3px & 5px & 8px & 12px & mean  \\
\hline\hline
FPN & \multirow{2}{*}{PointRend~\cite{kirillov2020pointrend}} & 46.9 & 53.6 & 59.3 & 63.8 & 55.9 \\ 
\ourmethod{} & \multirow{2}{*}{R50} & 49.2 & 56.2 & 62.0 & 66.4 & 58.5\\ 
\emph{improvement} & & \cellcolor{lightgray!60}{\textbf{(\emph{+2.3})}} & \cellcolor{lightgray!60}{\textbf{(\emph{+2.6})}} & \cellcolor{lightgray!60}{\textbf{(\emph{+2.7})}} & \cellcolor{lightgray!60}{\textbf{(\emph{+2.6})}} & \cellcolor{lightgray!60}{\textbf{(\emph{+2.6})}}  \\   
\hline\hline
FPN & \multirow{2}{*}{PointRend~\cite{kirillov2020pointrend}} & 47.8 & 54.6 & 60.5 & 64.9 & 57.0 \\ 
\ourmethod{} & \multirow{2}{*}{R101} & 50.1 & 57.1 & 62.9 & 67.2 & 59.3 \\ 
\emph{improvement} & & \cellcolor{lightgray!60}{\textbf{(\emph{+2.3})}} & \cellcolor{lightgray!60}{\textbf{(\emph{+2.5})}} & \cellcolor{lightgray!60}{\textbf{(\emph{+2.4})}} & \cellcolor{lightgray!60}{\textbf{(\emph{+2.3})}} & \cellcolor{lightgray!60}{\textbf{(\emph{+2.3})}}  \\\hline  
\end{tabular}%
}
\end{table}

\subsection{Main Results}
In this section, we present the detailed empirical comparisons to FPN \cite{lin2017feature} on four dense prediction tasks, including object detection, semantic, instance and panoptic segmentation in Table~\ref{tab:obj-det} - \ref{tab:pan-seg}, respectively. 

In general, \ourmethod{} substantially outperforms FPN on all scenarios of tasks and datasets. 
There are several detailed observations.
First, \ourmethod{} improves the primary evaluation metrics by \textbf{1.2 - 2.6 points} over FPN on all four tasks with ResNet50~\cite{he2016deep} as the bottom-up backbone. 
Second, the improvements brought by \ourmethod{} hold for stronger bottom-top backbones (e.g. ResNet101 \cite{he2016deep}) with a longer training schedule of 270K iterations. 
Third, the improvement from \ourmethod{} extends to more sophisticated mask heads, \eg PointRend \cite{kirillov2020pointrend}, on instance segmentation, as shown in Table~\ref{tab:ins-seg} (bottom section).

\begin{figure*}
\begin{center}
\includegraphics[width=0.98\linewidth,trim={3mm 0 0 0},clip]{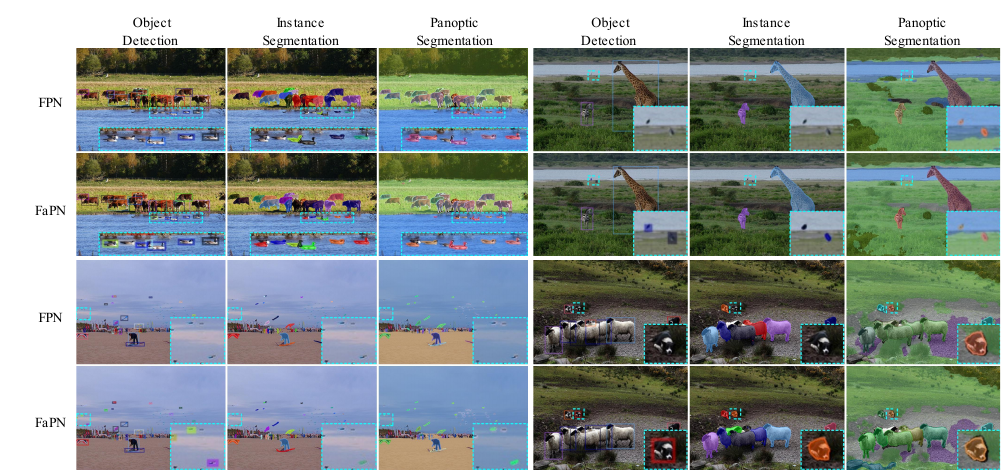}
\end{center}
   \caption{\textbf{Example pairs of results comparing FPN \cite{lin2017feature} and our \ourmethod{}}. Images are randomly chosen from \cite{lin2014microsoft}. Best view in color and zoom in for details.}
\label{fig:overview_results}
% \vspace{-0.5cm}
\end{figure*}

\begin{table}[!ht]
\centering
\caption{\textbf{Object Detection:} Performance comparisons on MS COCO \emph{val} set between FPN and \ourmethod{}.\label{tab:obj-det}}
\resizebox{0.45\textwidth}{!}{%
\begin{tabular}{@{}|l|c|cccc|@{}}
\hline
method & backbone & AP$^{bb}$ & AP$_{s}^{bb}$ & AP$_{m}^{bb}$ & AP$_{l}^{bb}$ \\ \hline\hline
FPN &  \multirow{2}{*}{Faster R-CNN~\cite{ren2015faster}} & 37.9 & 22.4 & 41.1 & 49.1 \\
\ourmethod{} (ours) & \multirow{2}{*}{R50} & 39.2 & 24.5 & 43.3 & 49.1 \\
\emph{improvement} &  & \cellcolor{lightgray!60}{\textbf{(\emph{+1.3})}} & \cellcolor{lightgray!60}{\textbf{(\emph{+2.1})}} & \cellcolor{lightgray!60}{\textbf{(\emph{+2.2})}} & \cellcolor{lightgray!60}{{(\emph{+0.0})}} \\ \hline\hline
FPN & \multirow{2}{*}{Faster R-CNN~\cite{ren2015faster}} & 42.0 & 25.2 & 45.6 & 54.6 \\
\ourmethod{} (ours) & \multirow{2}{*}{R101} & 42.8 & 27.0 & 46.2 & 54.9 \\
\emph{improvement} &  & \cellcolor{lightgray!60}{\textbf{(\emph{+0.8})}} & \cellcolor{lightgray!60}{\textbf{(\emph{+1.8})}} & \cellcolor{lightgray!60}{\textbf{(\emph{+0.6})}} & \cellcolor{lightgray!60}{\textbf{(\emph{+0.3})}}\\\hline
\end{tabular}%
}
\end{table}

\begin{table}[!ht]
\centering
\caption{\textbf{Semantic Segmentation:} Performance comparisons on Cityscapes \emph{val} set between FPN and \ourmethod{}.\label{tab:sem-seg}}
\resizebox{.45\textwidth}{!}{%
\begin{tabular}{@{}|l|c|cccc|@{}}
\hline
method & backbone & mIoU & iIoU & IoU\_sup & iIoU\_sup \\ \hline\hline
FPN & \multirow{2}{*}{PointRend~\cite{kirillov2020pointrend}} & 77.4 & 58.5 & 89.9 & 76.9 \\
\ourmethod{} (ours) & \multirow{2}{*}{R50} & 80.0 & 61.3 & 90.6 & 78.5 \\
\textit{improvement} &  & \cellcolor{lightgray!60}{\textbf{(\emph{+2.6})}} & \cellcolor{lightgray!60}{\textbf{(\emph{+2.8})}} & \cellcolor{lightgray!60}{\textbf{(\emph{+0.7})}} & \cellcolor{lightgray!60}{\textbf{(\emph{+1.6})}} \\ \hline\hline
FPN & \multirow{2}{*}{PointRend~\cite{kirillov2020pointrend}} & 78.9 & 59.9 & 90.4 & 77.8 \\
\ourmethod{} (ours) & \multirow{2}{*}{R101} & 80.1 & 62.2 & 90.8 & 78.6 \\
\textit{improvement} &  & \cellcolor{lightgray!60}{\textbf{(\emph{+1.2})}} & \cellcolor{lightgray!60}{\textbf{(\emph{+2.3})}} & \cellcolor{lightgray!60}{\textbf{(\emph{+0.4})}} & \cellcolor{lightgray!60}{\textbf{(\emph{+0.8})}} \\\hline
\end{tabular}%
}
% \vspace{-0.5cm}
\end{table}

\begin{table}[!hbt]
\centering
\caption{\textbf{Instance Segmentation:} Performance comparisons on MS COCO \emph{val} set between FPN and \ourmethod{}.
% \vspace{-0.2cm}
\label{tab:ins-seg}}
\resizebox{.48\textwidth}{!}{%
\begin{tabular}{@{}|l|c|cc|cc|@{}}
\hline
method & backbone & AP$^{mask}$ & AP$^{mask}_{s}$ & AP$^{bb}$ & AP$^{bb}_{s}$ \\ \hline\hline
FPN & \multirow{2}{*}{Mask R-CNN~\cite{he2017mask}} & 35.2 & 17.1 & 38.6 & 22.5 \\
\ourmethod{} (ours) & \multirow{2}{*}{R50}  & 36.4 & 18.1 & 39.8 & 24.3 \\
\emph{improvement} &  & \cellcolor{lightgray!60}{(\textit{\textbf{+1.2}})} & \cellcolor{lightgray!60}{(\textit{\textbf{+1.0}})} & \cellcolor{lightgray!60}{(\textit{\textbf{+1.2}})} & \cellcolor{lightgray!60}{(\textit{\textbf{+1.8}})}\\ 
\hline\hline
FPN & \multirow{2}{*}{Mask R-CNN~\cite{he2017mask}} & 38.6 & 19.5 & 42.9 & 26.4 \\
\ourmethod{} (ours) &  \multirow{2}{*}{R101} & 39.6 & 20.9 & 43.8 & 27.4 \\
\emph{improvement} &  & \cellcolor{lightgray!60}{(\textit{\textbf{+1.0)}}} & \cellcolor{lightgray!60}{(\textit{\textbf{+1.4}})} & \cellcolor{lightgray!60}{(\textit{\textbf{+0.9}})} & \cellcolor{lightgray!60}{(\textit{\textbf{+1.0}})} \\ 
\hline\hline
FPN & \multirow{2}{*}{PointRend~\cite{kirillov2020pointrend}} & 36.2 & 17.1 & 38.3 & 22.3 \\
\ourmethod{} + PR (ours) &  \multirow{2}{*}{R50} & 37.6 & 18.6 & 39.4 & 24.2 \\
\emph{improvement} &  & \cellcolor{lightgray!60}{(\textit{\textbf{+1.4}})} & \cellcolor{lightgray!60}{(\textit{\textbf{+1.5}})} & \cellcolor{lightgray!60}{(\textit{\textbf{+1.1}})} & \cellcolor{lightgray!60}{(\textit{\textbf{+1.9}})} \\\hline
\end{tabular}%
}
% \vspace{-0.4cm}
\end{table}

\begin{table}[!hbt]
\centering
\caption{\textbf{Panoptic Segmentation:} Performance comparisons on MS COCO \emph{val} set between FPN and \ourmethod{}.\label{tab:pan-seg}}
% \vspace{-0.2cm}
\resizebox{.49\textwidth}{!}{%
\begin{tabular}{@{}|l|c|c|cc|cc|@{}}
\hline
method & backbone & PQ & mIoU & PQ$^{St}$ & AP$^{bb}$ & PQ$^{Th}$ \\ \hline\hline
FPN & \multirow{2}{*}{PanopticFPN~\cite{kirillov2019panoptic}} & 39.4 & 41.2 & 29.5 & 37.6 & 45.9 \\ 
\ourmethod{} (ours) & \multirow{2}{*}{R50} & 41.1 & 43.4 & 32.5 & 38.7 & 46.9 \\
\textit{improvement} &  & \cellcolor{lightgray!60}{(\textit{\textbf{+1.7}})} & \cellcolor{lightgray!60}{(\textit{\textbf{+2.2}})} & \cellcolor{lightgray!60}{(\textit{\textbf{+3.0}})} & \cellcolor{lightgray!60}{(\textit{\textbf{+0.9}})} & \cellcolor{lightgray!60}{(\textit{\textbf{+1.0}})} \\ 
\hline\hline
FPN & \multirow{2}{*}{PanopticFPN~\cite{kirillov2019panoptic}} & 43.0 & 44.5 & 32.9 & 42.4 & 49.7 \\
\ourmethod{} (ours) & \multirow{2}{*}{R101} & 44.2 & 45.7 & 35.0 & 43.0 & 50.3 \\
\textit{improvement} &  & \cellcolor{lightgray!60}{(\textit{\textbf{+1.2}})} & \cellcolor{lightgray!60}{(\textit{\textbf{+1.2}})} & \cellcolor{lightgray!60}{(\textit{\textbf{+2.1}})} & \cellcolor{lightgray!60}{(\textit{\textbf{+0.6}})} & \cellcolor{lightgray!60}{(\textit{\textbf{+0.6}})} \\
\hline\hline
FPN & \multirow{2}{*}{PanopticFCN~\cite{li2021fully}} & 41.1 & 79.8 & 49.9 & 30.2 & 41.4 \\
\ourmethod{} (ours) & \multirow{2}{*}{R50} & 41.8 & 80.2 & 50.5 & 30.8 & 42.0 \\
\textit{improvement} &  & \cellcolor{lightgray!60}{(\textit{\textbf{+0.7}})} & \cellcolor{lightgray!60}{(\textit{\textbf{+0.4}})} & \cellcolor{lightgray!60}{(\textit{\textbf{+0.6}})} & \cellcolor{lightgray!60}{(\textit{\textbf{+0.6}})} & \cellcolor{lightgray!60}{(\textit{\textbf{+0.6}})} \\
\hline\hline
FPN & \multirow{2}{*}{PanopticFCN~\cite{li2021fully}} & 42.7 & 80.8 & 51.4 & 31.6 & 43.9 \\
\ourmethod{} (ours) & \multirow{2}{*}{R50-600} & 43.5 & 81.3 & 52.1 & 32.3 & 53.5 \\
\textit{improvement} &  & \cellcolor{lightgray!60}{(\textit{\textbf{+0.8}})} & \cellcolor{lightgray!60}{(\textit{\textbf{+0.5}})} & \cellcolor{lightgray!60}{(\textit{\textbf{+0.7}})} & \cellcolor{lightgray!60}{(\textit{\textbf{+0.7}})} & \cellcolor{lightgray!60}{(\textit{\textbf{+0.6}})} \\\hline
\end{tabular}%
}
\vspace{-0.2cm}
\end{table}

In particular, we notice that the improvement is larger on small objects (e.g. AP$^{bb}_{s}$, AP$^{mask}_{s}$). 
For instance, \ourmethod{} improves the bounding box AP on small objects by \textbf{2.1 points} and \textbf{1.8 points} over FPN on MS COCO object detection and instance segmentation, respectively. Conceptually, small objects occupy fewer pixels in an image, and most of pixels are distributed along the object boundaries. Hence, it is vital to be able to correctly classify the boundaries for small objects. 
However, as features traverse the top-bottom pathway through heuristics-based upsampling operations (\eg FPN uses nearest neighbor upsampling), shifts in pixels (\ie misalignment) are foreseeable and the amount of shifts will accumulate as the number of upsampling steps increases. 
Hereby, the severity of the misalignment will reach its maximum at the finest feature maps in the top-down pathway pyramid, which are typically used for detecting or segmenting small objects, resulting in a significant degradation in performance. 
On the other hand, \ourmethod{} performs feature alignment progressively which in turn alleviates the misalignment at the finest level step by step, and thus achieves significant improvements on small objects compared to the FPN~\cite{lin2017feature}. Qualitative improvements are also evidenced in Figure~\ref{fig:overview_results}. Finally, we incorporate \ourmethod{} into MaskFormer~\cite{cheng2021per}, and demonstrate that \ourmethod{} leads to state-of-the-art performance on two complex semantic segmentation tasks, \ie ADE20K and COCO-Stuff-10K, as shown in Table~\ref{tab:sota}. 

\begin{table}[!ht]
\centering
\caption{\textbf{Comparison to SOTA} on (a) ADE20K \texttt{val} and (b) COCO-Stuff-10K \texttt{test}. We report both single-scale (s.s.) and multi-scale (m.s.) semantic segmentation performance. Backbones pre-trained on ImageNet-22K are marked with $^{\text{\textdagger}}$. Our results are highlighted in shade.\label{tab:sota}}
\vspace{-.8em}
\begin{subtable}[h]{0.48\textwidth}
\caption{{\footnotesize ADE20K \texttt{val}}\label{tab:sem-seg-ade20k}}
\vspace{-.3em}
\resizebox{\textwidth}{!}{%
\begin{tabular}{@{}|l|c|c|cc|@{}}
\hline
method & backbone & crop size & mIoU (s.s.) & mIoU (m.s.)\\ \hline\hline
OCRNet~\cite{yuan2019object} & R101 & $520\times520$ & - & 45.3 \\
AlignSeg~\cite{huang2020alignseg} & R101 & $512\times512$ & - & 46.0 \\
SETR~\cite{zheng2021rethinking} & ViT-L$^{\text{\textdagger}}$ & $512\times512$ & - & 50.3 \\
Swin-UperNet~\cite{liu2021swin} & Swin-L$^{\text{\textdagger}}$ & $640\times640$ & - & 53.5 \\
MaskFormer~\cite{cheng2021per} & Swin-L$^{\text{\textdagger}}$ & $640\times640$ & 54.1 & 55.6 \\
\rowcolor{lightgray!60}
MaskFormer + FaPN & Swin-L$^{\text{\textdagger}}$ & $640\times640$ & \textbf{55.2} & \textbf{56.7}\\\hline
\end{tabular}%
}
\end{subtable}\\
\vspace{1em}
\begin{subtable}[h]{0.48\textwidth}
\caption{\footnotesize COCO-Stuff-10K \texttt{test}\label{tab:sem-seg-coco-v1}}
\vspace{-.3em}
\resizebox{\textwidth}{!}{%
\begin{tabular}{@{}|l|c|c|cc|@{}}
\hline
method & backbone & crop size & mIoU (s.s.) & mIoU (m.s.)\\ \hline\hline
OCRNet~\cite{yuan2019object} & \multirow{3}{*}{R101} & $520\times520$ & - & 39.5 \\
MaskFormer~\cite{cheng2021per} &  & $640\times640$ & 38.1 & 39.8 \\
\rowcolor{lightgray!60}
MaskFormer + FaPN &  & $640\times640$ & \textbf{39.6} & \textbf{40.6}\\\hline
\end{tabular}%
}
\end{subtable}
\end{table}

Overall, extensive comparisons on scenarios comprised of different tasks and datasets have confirmed the effectiveness of our proposed \ourmethod{} for dense image prediction. 
A straightforward replacement of FPN with \ourmethod{} achieves substantial improvements without bells and whistles. The generality and flexibility to different bottom-up backbones or mask heads have further strengthened the practical utilities of \ourmethod{}. 

\subsection{Real-time Performance}
Driven by real-world applications (\eg, autonomous driving), there are growing interests in real-time dense prediction, which requires the generation of high-quality predictions with minimal latency. In this section, we aim to investigate the effectiveness of our proposed \ourmethod{} under real-time settings, \ie inference speed $\geq$ 30 FPS. The full details are provided in the supplementary materials. 

We compare our \ourmethod{} with state-of-the-art real-time semantic segmentation methods on Cityscapes and COCO-Stuff-10K in Table~\ref{tab:realtime-sem-seg}, in terms of accuracy (mIoU) and inference speed (FPS). In general, we observe that a straightforward replacement of FPN with the proposed \ourmethod{} results in a competitive baseline against other dedicated real-time semantic segmentation methods. 

In particular, on Cityspaces, \ourmethod{}-R18 runs 2$\times$ faster than SwiftNet~\cite{orsic2019defense}, while maintaining a similar mIoU performance. In addition, with a larger backbone and input size, \ourmethod{}-R34 achieves a competitive mIoU of 78.1 points on the \emph{test} split, in the same time outputting 30 FPS. On the more challenging COCO-Stuff-10K, our \ourmethod{} also outperforms other existing methods by a substantial margin. Specifically, \ourmethod{}-R34 outperforms BiSeNetV2~\cite{yu2020bisenet} in both segmentation accuracy measured in mIoU and inference speed.

\begin{table}[!ht]
\centering
\caption{\textbf{Real-time semantic segmentation} on (a) Cityscapes and (b) COCO-Stuff-10K. $\dagger$ denotes a method with a customized backbone. Our results are highlighted in shade.\label{tab:realtime-sem-seg}}
\vspace{-1em}
\begin{subtable}[h]{0.48\textwidth}
    \caption{{\footnotesize Cityscapes}}
    \vspace{-.3em}
    \resizebox{\textwidth}{!}{%
    \begin{tabular}{@{}|l|c|c|ccc|@{}}
    \hline
    method & backbone & crop size & FPS & mIoU (\emph{val}) & mIoU (\emph{test}) \\ \hline\hline
    ESPNet~\cite{mehta2018espnet} & $\dagger$ & {$512\times1024$} & 113 & - & 60.3 \\
    ESPNetV2~\cite{mehta2019espnetv2} & $\dagger$ & {$512\times1024$} & - & 66.4 & 66.2 \\
    \rowcolor{lightgray!40}
    FaPN & R18 & {$512\times1024$} & \textbf{142} & \textbf{69.2} & \textbf{68.8} \\ \hline\hline
    BiSeNet~\cite{yu2018bisenet} & {R18} & {$768\times1536$} & 65.6 & 74.8 & 74.7 \\
    \rowcolor{lightgray!50}
    FaPN & {R18} & {$768\times1536$} & \textbf{78.1} & \textbf{75.6} & \textbf{75.0} \\ \hline\hline
    SwiftNet~\cite{orsic2019defense} & R18 & {$1024\times2048$} & \textbf{39.9} & 75.4 & 75.5 \\
    ICNet~\cite{zhao2018icnet} & R50 & {$1024\times2048$} & 30.3 & - & 69.5 \\
    \rowcolor{lightgray!60}
    FAPN & R34 & {$1024\times2048$} & 30.2 & \textbf{78.5} & \textbf{78.1} \\\hline
    \end{tabular}%
    }
\end{subtable}\\
\vspace{1em}
\begin{subtable}[h]{0.39\textwidth}
    \caption{{\footnotesize COCO-Stuff-10K}}
    \vspace{-.3em}
    \resizebox{\textwidth}{!}{%
    \begin{tabular}{@{}|l|c|c|cc|@{}}
    \hline
    method & backbone & crop size & FPS & mIoU (\emph{val}) \\ \hline\hline
    BiSeNet~\cite{yu2018bisenet} & R18 & \multirow{5}{*}{$640\times640$} & - & 28.1 \\
    BiSeNetV2~\cite{yu2020bisenet} & $\dagger$ &  & 42.5 & 28.7 \\
    ICNet~\cite{zhao2018icnet} & R50 &  & 35.7 & 29.1 \\
    \rowcolor{lightgray!40}
    FaPN & R18 &  & \textbf{154} & 28.4 \\
    \rowcolor{lightgray!60}
    FaPN & R34 &  & 110 & \textbf{30.3} \\\hline
    \end{tabular}%
    }
\end{subtable}
\end{table}

\section{Conclusion}    \label{conclusion}
This paper introduced \emph{Feature-aligned Pyramid Network (\ourmethod{})}, a simple yet effective top-down pyramidal architecture to generate multi-scale features for dense image prediction. It is comprised of a feature alignment module that learns transformation offsets of pixels to contextually align upsampled higher-level features; and a feature selection module to emphasize the lower-level features with rich spatial details. Empirically, \ourmethod{} leads to substantial and consistent improvements over the original FPN on four dense prediction tasks and three datasets. Furthermore, \ourmethod{} improves the state-of-the-art segmentation performance when integrated in strong baselines. Additionally, \ourmethod{} can be easily extended to real-time segmentation tasks by pairing it with lightweight backbones, where we demonstrate that \ourmethod{} performs favorably against dedicated real-time methods. In short, given the promising performance on top of a simple implementation, we believe that \ourmethod{} can serve as the new baseline/module for dense image prediction. 

\paragraph{Acknowledgments:} This work was supported by the National Natural Science Foundation of China (Grant No. 61903178, 61906081, and U20A20306) and the Program for Guangdong Introducing Innovative and Enterpreneurial Teams (Grant No. 2017ZT07X386).

\newpage
\appendix
\begin{center}{\bf \Large Appendix}\end{center}\vspace{-2mm}
\renewcommand{\thetable}{\Roman{table}}
\renewcommand{\thefigure}{\Roman{figure}}
\setcounter{table}{0}
\setcounter{figure}{0}
In this supplementary material we include (1) additional training details in Section~\ref{sec:app_train}; (2) more details on the real-time semantic segmentation experiment in Section~\ref{sec:app_real}; and (3) additional qualitative visualizations to demonstrate the effectiveness of the proposed \ourmethod{} in Section~\ref{sec:app_vis}. 

\section{Training Settings\label{sec:app_train}}
For all experiments shown in the main paper, we use SGD optimizer with 0.9 momentum and 0.0001 weight decay. 
The standard data augmentation of horizontal flipping and scaling are also applied. 
In addition, the weights of the batch normalization~\cite{ioffe2015batch} layers derived from the ImageNet pre-trained models are kept frozen. 
To be consistent with prior works, we have not incorporated any testing time augmentation tricks. 
For semantic segmentation, the model is trained for 65K iterations starting with a learning rate of 0.01 that is reduced by a factor of 10 at 40K and 55K. 
For the other three dense prediction tasks, the model is trained for 90K or 270K iterations with the initial learning rate of 0.02 that is reduced to 0.002 at 60K/210K and 0.0002 at 80K/250K. 
Our implementation is based on the Detectron2~\cite{wu2019detectron2} with the default configurations, \ie to maintain a fair comparison with prior works, neither have we tuned any training hyperparameters nor used advanced data augmentations. 

\section{Real-time Semantic Segmentation Continued\label{sec:app_real}}
With a lightweight ResNet (\eg ResNet18/34) as the bottom-up backbone, we denote the feature maps output by the last three stages, \ie conv3, conv4, conv5, as $\left\{\mathbf{C}_{3}, \mathbf{C}_{4}, \mathbf{C}_{5}\right\}$, respectively.  
At the beginning, we simply attach an FSM layer on $\mathbf{C}_{5}$ to produce the coarsest resolution feature maps $\mathbf{P}_{5} \in \mathbb{R}^{128 \times \frac{H}{32} \times \frac{W}{32}} $ (\ie the output channel of FSM is 128).
With a coarser-resolution feature map $\mathbf{P}_{i}$ ($l \in [4, 5]$), we upsample its spatial resolution by a factor of 2 using nearest neighbor upsampling~\cite{lin2017feature} to obtain $\mathbf{P}_{i}^{up} \in \mathbb{R}^{128 \times \frac{H}{2^{i-1}} \times \frac{W}{2^{i-1}}} $.
Afterwards, an FAM layer is used to align $\mathbf{P}_{i}^{up}$ to its corresponding bottom-up feature map $\mathbf{\hat{C}}_{i-1} \in \mathbb{R}^{128 \times \frac{H}{2^{i-1}} \times \frac{W}{2^{i-1}}} $ deriving from $\mathbf{C}_{i-1}$ by undergoing an FSM layer for channel reduction. 
Instead of element-wise addition, the aligned $\mathbf{\hat{P}}_{i}^{up}$ is then merged with $\mathbf{\hat{C}}_{i-1}$ by concatenation along with channel, and the merged feature map $\mathbf{P}_{i-1} \in \mathbb{R}^{256 \times \frac{H}{2^{i-1}} \times \frac{W}{2^{i-1}}}$ which has the identical spatial size to $\mathbf{C}_{i-1}$ is further input in a Conv $1\times1$ layer to reduce its channels to 128.
This process is iterated until the finest resolution map $\mathbf{P}_{3} \in \mathbb{R}^{128 \times \frac{H}{8} \times \frac{W}{8}} $ is generated.
Finally, we append a prediction layer on $\mathbf{P}_{3}$ to generate the final semantic mask.

We train our models using the SGD optimizer with momentum and weight decay set to 0.9 and 0.0005, respectively.
During training, we apply random horizontal flip and scale to input images, followed by a crop to a fixed size.
The scale is randomly selected from  $\left\{0.75, 1, 1.25, 1.5, 1.75, 2.0\right\}$, and the cropped resolutions are $1536\times768$ and $640\times640$ for Cityscapes~\cite{cordts2016cityscapes} and COCO-Stuff \cite{caesar2018coco}, respectively. 
For all datasets, we set the batch size and the initial learning rate to 16 and 0.01, while the learning rate decays following a ``poly" strategy, specifically $0.01 \times (1-\frac{iter}{maxiters})^{0.9}$.  
Following the prior works~\cite{yu2018bisenet,yu2020bisenet}, we train models for 40K and 20K training iterations on Cityscapes and COCO-Stuff, respectively. 
In the evaluation process, we compute the inference speed using one Titan RTX GPU without any speed-up trick (\eg, Apex or TensorRT) or optimized depth-wise convolutions, and evaluate the accuracy without any testing augmentation technique (\eg, multi-crop or multi-scale testing). 

\begin{table}[!ht]
\centering
\caption{\textbf{Ablation Study on Real-time Semantic Segmentation:} Detailed comparisons of each component in our proposed real-time semantic segmentation \ourmethod{} over Citsycapes \emph{val} set in terms of accuracy, parameters and FLOPs (computational complexity). \label{tab:abl_realtime-sem-seg}}
\resizebox{.45\textwidth}{!}{%
\begin{tabular}{|l|c|c|c|}
\hline
method & mIoU & \#Params (M) & FLOPs (G) \\\hline\hline
FPN & 68.6 & 11.4 & 44.5 \\
+ FAM    & 73.8 & 12.2 & 51.0 \\ 
+ FAM + FSM & 74.2 & 12.6 & 51.0 \\
\rowcolor{lightgray!60}
+ FAM + FSM + Concat & 75.6 & 12.6 & 51.8 \\\hline
% \bottomrule
\end{tabular}%
}
\end{table}

We first demonstrate the effectiveness of each module in our proposed real-time semantic segmentation framework separately and then study different feature fusion methods on the Cityscapes \emph{val} set. We use ResNet18 pre-trained on ImageNet as our backbone in the following ablation analysis. 

The ablation experimental results are given in Table~\ref{tab:abl_realtime-sem-seg}.
Basically, the incorporation of FAM into the baseline improves the performance from 68.6\% to 73.8\%.
Furthermore, FSM improves the performance to 74.2\% with only 0.4$M$ additional parameters.
Besides, when we replace the element-wise sum operation with concatenation for fusing the detailed feature and aligned semantic feature, another 1.4\% improvement is achieved with few extra FLOPs. 

\begin{figure*}[!ht]
     \centering
     \includegraphics[width=1\linewidth]{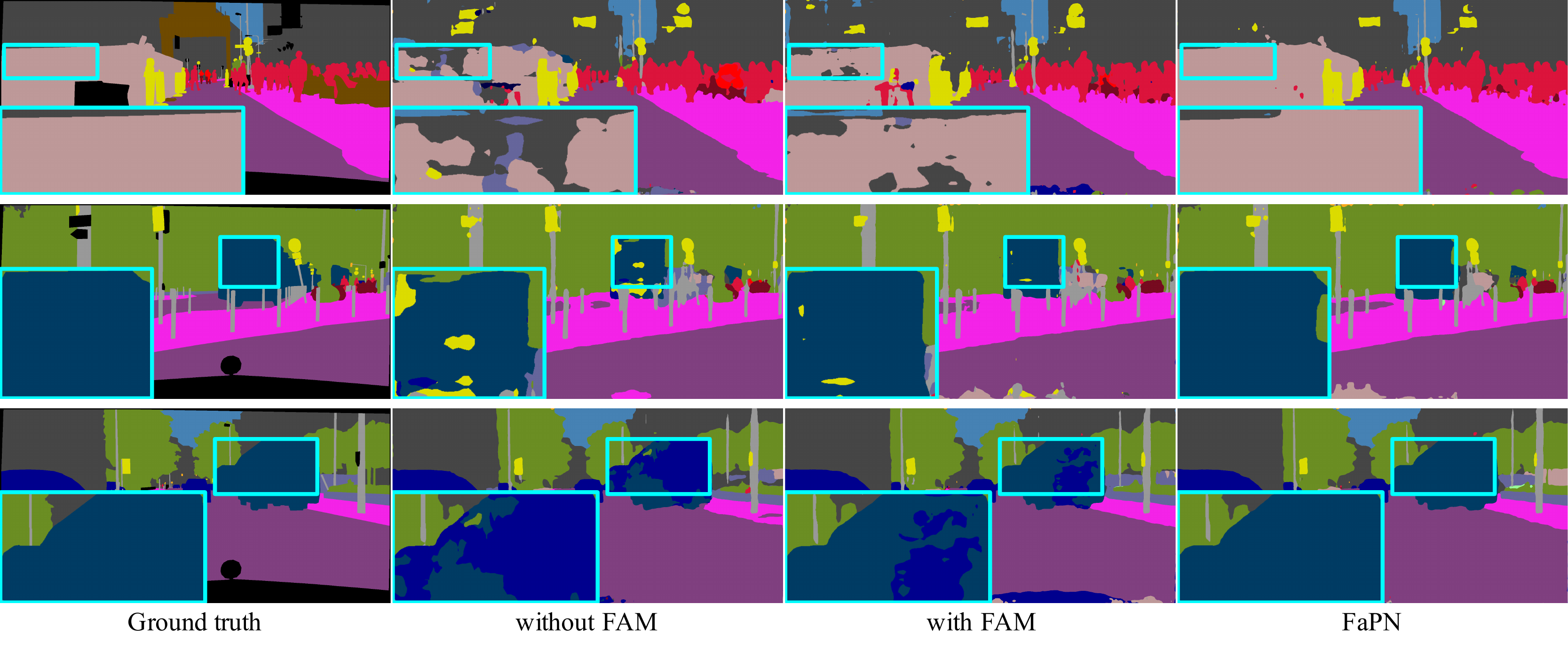}
     \caption{\textbf{Example visual comparisons among different approaches on the Cityscapes \emph{val} set}. From left to right: ground truths, results from baseline, baseline with FAM and our \ourmethod{}, respectively. All models are using ResNet18.}\label{fig:ab_FaPN}
\end{figure*}

Figure~\ref{fig:ab_FaPN} visualizes the semantic segmentation results of \ourmethod{} on Cityscapes under real-time settings (FPS $\geq$ 30)\footnote{\url{https://paperswithcode.com/sota/real-time-semantic-segmentation-on-cityscapes}}.
Noticeably, the proposed Feature Align Module (FAM; third column in Figure~\ref{fig:ab_FaPN}) significantly improves the segmentation quality from the baseline (\ie FPN; second column in Figure~\ref{fig:ab_FaPN}). With the feature selection module and feature concatenation, our final approach \ourmethod{} further improves the performance on real-time semantic segmentation.

\section{Additional Visualization\label{sec:app_vis}}
\begin{figure*}[!ht]
     \centering
     \includegraphics[width=0.95\linewidth]{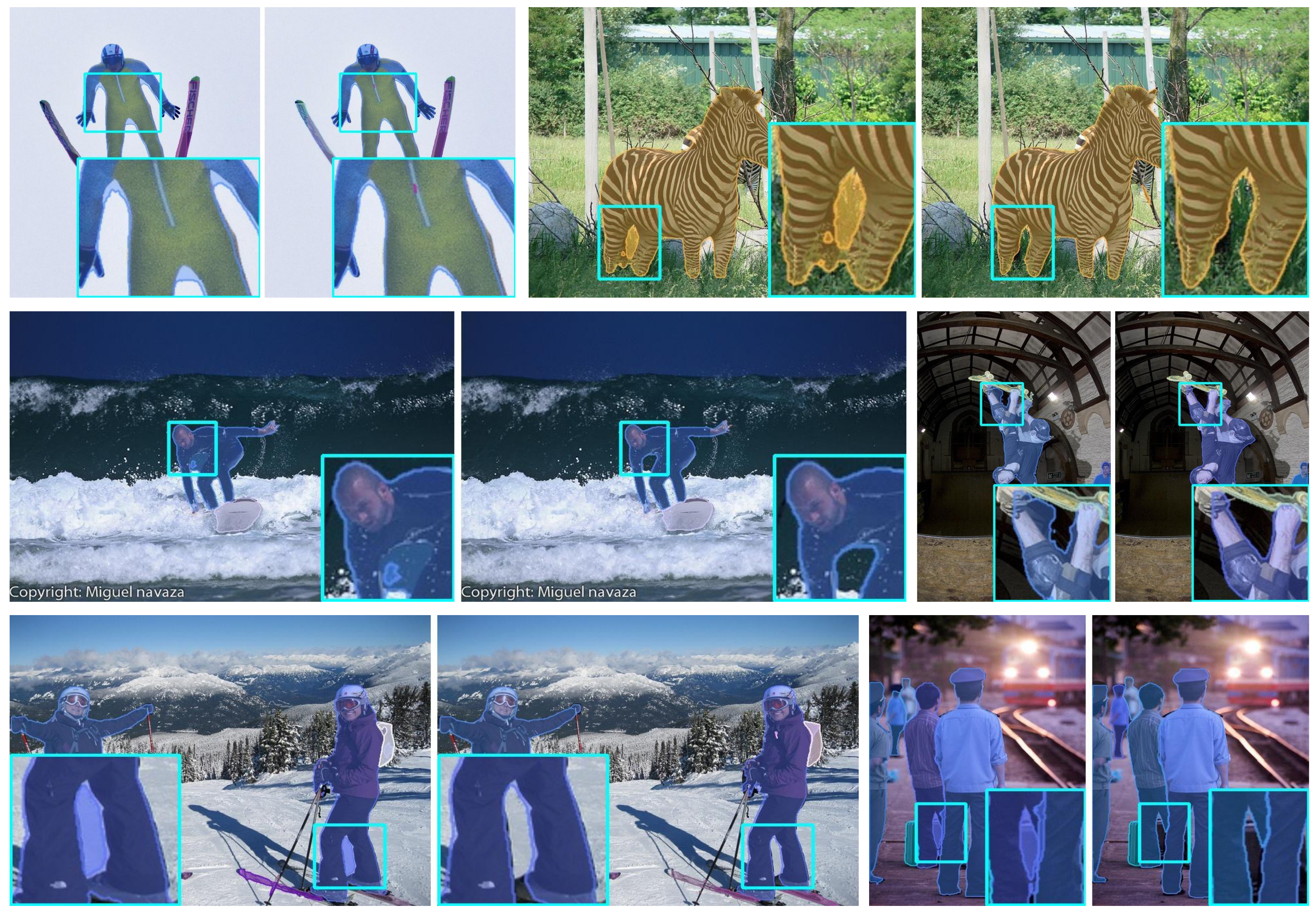}
     \caption{\label{fig:add_boundaries} \textbf{Example pairs of instance segmentation results from FPN (\emph{Left}) and our \ourmethod{} (\emph{Right}) on object boundaries}. Both methods are implemented in Mask R-CNN with ResNet-50 being the backbone and PointRend as the mask head.}
\end{figure*}

\begin{figure*}[!ht]
     \centering
     \includegraphics[width=0.95\linewidth]{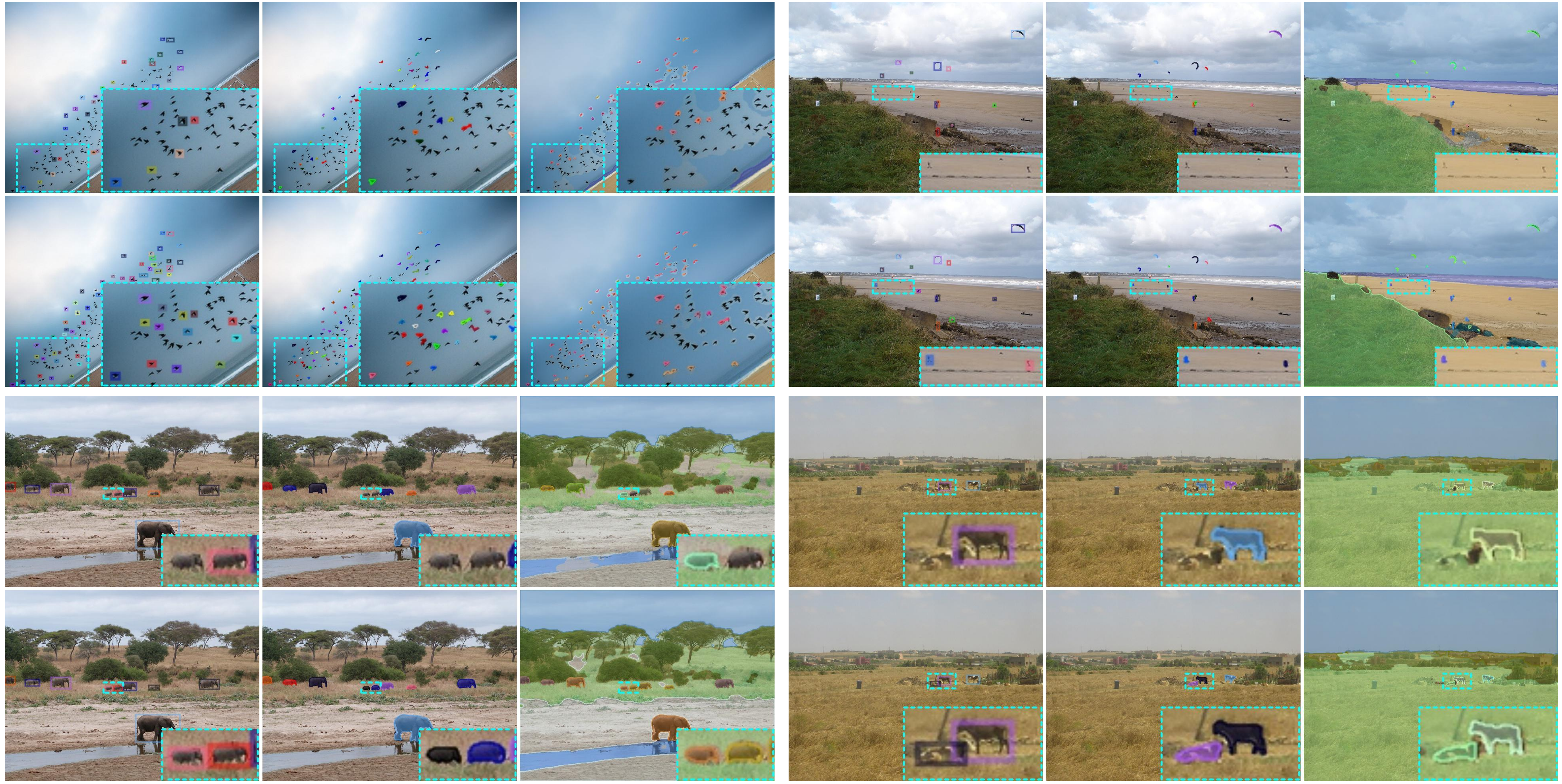}
     \caption{\label{fig:add_small} \textbf{Example results from methods with FPN or our \ourmethod{} on small objects}. In each group, from left to right, they are object detection, instance and panoptic segmentation, the top is achieved by FPN while the bottom by our \ourmethod{}. All models are based on ResNet50.}  PointRend head
\end{figure*}

Figure~\ref{fig:add_boundaries} and Figure~\ref{fig:add_small} visualize the dense prediction performance on MS COCO. Evidently, our method achieves a more accurate segmentation on object boundaries and small objects. 

{\small
\bibliographystyle{ieee_fullname}
\bibliography{egbib}
}

\end{document}